\documentclass[letterpaper, 10 pt, conference]{ieeeconf}  

\IEEEoverridecommandlockouts                              

\usepackage{color}
\definecolor{cvprblue}{rgb}{0.21,0.49,0.74}
\usepackage[pagebackref,breaklinks,colorlinks,citecolor=cvprblue]{hyperref}

\usepackage{epsfig}
\usepackage{graphicx}
\usepackage{amsmath}
\usepackage{amssymb}
\usepackage[ruled,linesnumbered]{algorithm2e}
\usepackage{subcaption}
\usepackage{booktabs} 
\usepackage{bm}
\usepackage{multirow}
\usepackage{cite}
\usepackage{hyperref}
\definecolor{gray}{RGB}{120,120,120}

\title{\LARGE \bf
SAM-PD: How Far Can SAM Take Us in Tracking and Segmenting Anything in Videos by Prompt Denoising
}

\author{Tao Zhou$^1$ \quad Wenhan Luo$^{2}$ \quad Qi Ye$^{1}$ \quad Zhiguo Shi$^1$ \quad Jiming Chen$^1$
\thanks{$^1$Zhejiang University, {\tt\small zhoutao2015@zju.edu.cn}}
\thanks{$^2$Hong Kong University of Science and Technology}
}

\begin{document}
\maketitle
\thispagestyle{empty}
\pagestyle{empty}

\begin{abstract}

Recently, promptable segmentation models, such as the Segment Anything Model (SAM), have demonstrated robust zero-shot generalization capabilities on static images. These promptable models exhibit denoising abilities for imprecise prompt inputs, such as imprecise bounding boxes. In this paper, we explore the potential of applying SAM to track and segment objects in videos where we recognize the tracking task as a prompt denoising task. Specifically, we iteratively propagate the bounding box of each object's mask in the preceding frame as the prompt for the next frame. Furthermore, to enhance SAM's denoising capability against position and size variations, we propose a multi-prompt strategy where we provide multiple jittered and scaled box prompts for each object and preserve the mask prediction with the highest semantic similarity to the template mask. We also introduce a point-based refinement stage to handle occlusions and reduce cumulative errors. Without involving tracking modules, our approach demonstrates comparable performance in video object/instance segmentation tasks on three datasets: DAVIS2017, YouTubeVOS2018, and UVO, serving as a concise baseline and endowing SAM-based downstream applications with tracking capabilities. Code will be available at \href{https://github.com/infZhou/SAM-PD}{https://github.com/infZhou/SAM-PD}. 
\end{abstract}    
\section{Introduction}
\label{sec: intro}
Tracking and segmenting open-set objects is fundamental to approaching modern automation. Most existing video object segmentation (VOS) methods~\cite{SiamMask_cvpr19, Xmem_eccv22, DeAOT_nips22, STM_iccv19} propagate initial masks to subsequent frames via various model designs and have shown great effectiveness on closed-set data. Recently, facilitated by an exceptionally large dataset, a promptable image segmentation model, named Segment Anything Model (SAM)~\cite{SAM_iccv23}, has demonstrated a revolutionary improvement in zero-shot generalization capability. Several recent studies~\cite{SAM_PT_arxiv23, TAM_arxiv23, SAM_Track_arxiv23} have attempted to integrate SAM with other tracking modules for video segmentation. For instance, TAM~\cite{TAM_arxiv23} presents a tracking anything approach by combining SAM with an advanced mask tracker, XMem~\cite{Xmem_eccv22}. SAM-PT~\cite{SAM_PT_arxiv23} engages point trackers~\cite{PIPS_eccv22, Tap_Vid_nips22} to propagate points sampled from initial masks to subsequent frames, providing point prompts for SAM to obtain mask predictions.   

\begin{figure}[t]
  \centering
  \includegraphics[width=\columnwidth]{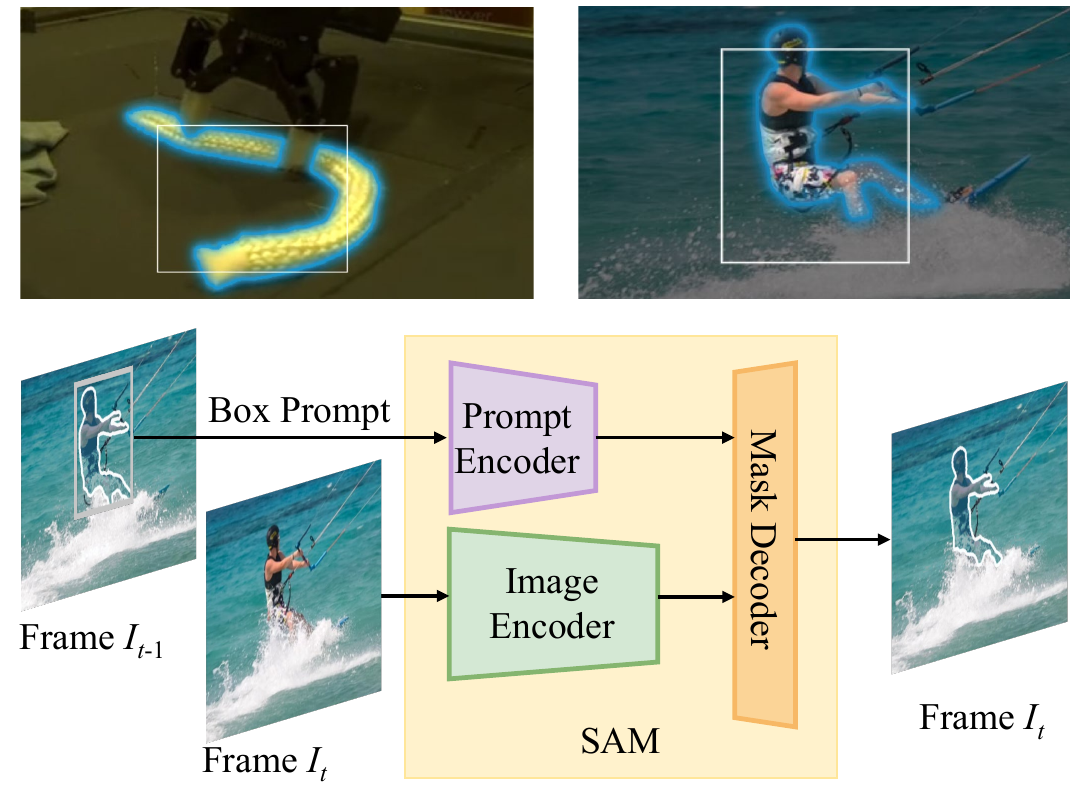}
  \caption{Top row: two examples where SAM works well with noisy bounding box prompts. Bottom row: our vanilla idea of utilizing the denoising capability of SAM to track and segment objects throughout a video.}
  \label{fig: fig_1}
\end{figure}

As illustrated in Fig.~\ref{fig: fig_1}, we observe that the promptable segmentation model, SAM~\cite{SAM_iccv23}, exhibits denoising capabilities for imprecise positional queries (e.g., imprecise bounding boxes). This capability arises from a compromise in interactive segmentation~\cite{SAM_iccv23, SEEM_nips23, interactive_segment1_arxiv20, interactive_segment2_icip20}, where a user may provide a loose box. During the training phase, SAM~\cite{SAM_iccv23} injects random noise into each box prompt, with a standard deviation of 10\% box side length (up to a maximum of $20$ pixels).

Driven by the above observations, in this paper, we report that by treating the tracking task as a prompt denoising task, SAM can be extended to segment objects in videos without incorporating external tracking modules, achieving comparable performance. Specifically, given a preceding frame and masks of objects, our vanilla idea propagates the bounding boxes of these masks to the next frame as box prompts for corresponding objects. Intrinsically, this assumes a slow movement of objects between adjacent frames.

Despite its effectiveness in simple cases, several challenges still exist. First, SAM's denoising capability undergoes reduced effectiveness when objects experience larger displacements and size variations. Second, SAM faces challenges in recovering a complete mask from over-tight box prompts, leading to cumulative errors, especially when objects are partially occluded. To address the first challenge, we propose a multi-prompt strategy. This consists of two steps. (1) In addition to the original box from the previous frame, we provide multiple jittered and scaled box prompt duplications for each object, enriching the coverage of different positions and scales, and yielding multiple mask predictions. (2) Inspired by the siamese framework~\cite{SiamFC_eccv16} in the tracking community, we compute the semantic similarity between the predicted mask and the template mask (in the first frame), preserving the predicted mask with the maximum semantic similarity. We extract semantic embeddings from the image embedding encoded by the SAM image encoder instead of involving additional encoding modules to reduce computational overhead.

To address the second challenge, we introduce a point-based mask refinement stage. The key insight is to leverage SAM's multiple prediction capability to recover complete masks covering all visible regions of objects. We sample positive point prompts from the coarse mask predicted via the box prompt and preserve the refined mask prediction with the maximum area (among multiple mask predictions). To avoid confusion among multiple targets, we attach points sampled from other objects as negative point prompts for the current object. The proposed point-based refinement demonstrates effectiveness in handling occlusions and mitigating cumulative errors. As our extensions do not require rerunning the heavy image encoder, the extra computational cost is acceptable, as analyzed in our experiments.

To summarize, our contributions are as follows:
\begin{itemize}
\item An SAM-based online method is proposed for zero-shot video object tracking and segmentation, where we treat the tracking task as a prompt denoising task. 

\item Two extensions are introduced to enhance SAM's capability in handling variations in object position, size, and visibility, mitigating cumulative errors.

\item The effectiveness is validated by extensive experiments on DAVIS2017~\cite{DAVIS_dataset_arxiv17}, YouTubeVOS2018~\cite{YouTube2018_dataset}, and UVO~\cite{UVO_dataset_ICCV21} datasets. SAM-PD serves as a concise baseline, endowing SAM-based downstream applications with tracking capabilities.
\end{itemize}

\section{Related Work}
\label{sec: Related Work}
\subsection{Box-level Detection and Tracking}
Due to the relatively low annotation cost, box-level detection~\cite{FRCNN_iccv15, yolo9000_cvpr17, CenterNet_arxiv19, DETR_eccv20} and tracking~\cite{SiamFC_eccv16, SiamRCNN_cvpr20, FairMOT_ijcv21, ByteTrack_eccv22, SORT_icip16, DeepSort_icip17, tracktor_cvpr19} have been studied for years, accumulating significant progress. In some frameworks, the task of object detection can be interpreted as a denoising task. For instance, the anchor-based detector, Faster R-CNN~\cite{FRCNN_iccv15} utilizes a regression head that absorbs predefined anchors and produces refined bounding boxes. Recently, Zhang et al.~\cite{DINO_iccv21, DN_DETR_cvpr22} interpreted the decoding process of learnable object queries in the end-to-end transformer-based detector DETR~\cite{DETR_eccv20} as a denoising task and improves the training speed of DETR by additionally feeding noisy ground-truth bounding boxes into the transformer decoder.

According to the number of objects to be tracked, tracking tasks can be classified into single-object tracking (SOT) and multi-object tracking (MOT). By adopting a fully convolutional siamese network, SiamFC~\cite{SiamFC_eccv16} set a representative framework for modern SOT. Furthermore, Siam R-CNN~\cite{SiamRCNN_cvpr20} introduces a re-detection mechanism, enabling improved effectiveness in the face of changes in object size and visibility. MOT~\cite{FairMOT_ijcv21, ByteTrack_eccv22, SORT_icip16, DeepSort_icip17, tracktor_cvpr19} aims to construct moving trajectories for number-agnostic objects. This requires the multi-object tracker to be capable of perceiving the birth, continuation, and termination of targets. To this end, most MOT methods follow the tracking-by-detection paradigm, working together with a detector. After a frame-by-frame detection, detections across frames are usually linked using motion cues~\cite{SORT_icip16, ByteTrack_eccv22} or appearance cues~\cite{FairMOT_ijcv21, DeepSort_icip17}. Related to our work, Tracktor++~\cite{tracktor_cvpr19} leverages the denoising capability of the detector's regression head~\cite{FRCNN_iccv15} and uses the tracked box as the proposal for the next frame, so that propagating the tracking identity, extending a detector to a multi-object tracker. In this paper, we aim to integrate these advancements with SAM, exploring its application for tracking and segmenting objects without introducing external tracking modules.

\subsection{Object Segmentation in Images and Videos}
Inspired by the success of pre-trained large language models, SAM~\cite{SAM_iccv23} aims to transfer such success to computer vision by building a foundation model for promptable image segmentation. Facilitated by the powerful data engine, the dataset built by SAM, named, SA-1B, includes more than 1B masks from 11M images and enables the trained model with revolutionary zero-shot generalization. Almost in parallel with SAM, another promptable segmentation model named SEEM~\cite{SEEM_nips23} works similarly and supports additional prompt modalities (e.g., text, audio, and sketches) compared to the released version of SAM.

In the task of video object segmentation, extensive efforts~\cite{STC_nips20, Xmem_eccv22, DeAOT_nips22, SiamMask_cvpr19} have been made in various aspects. For instance, SiamMask~\cite{SiamMask_cvpr19} developed a simple and fast-speed approach by augmenting the object-tracking framework SiamFC~\cite{SiamFC_eccv16} with a binary segmentation task. XMem~\cite{Xmem_eccv22} aimed at addressing the challenge of segmenting objects in long videos by proposing a unified memory architecture. STC~\cite{STC_nips20} proposed a self-supervised approach for learning representations for visual correspondence from raw videos, where videos are represented as graphs. Recently, we noticed some concurrent works~\cite{TAM_arxiv23, SAM_Track_arxiv23, SAM_PT_arxiv23} that combined SAM with powerful mask trackers~\cite{Xmem_eccv22, DeAOT_nips22} or point trackers~\cite{PIPS_eccv22, Tap_Vid_nips22} to conduct zero-shot video object segmentation. Diverging from these studies, in this paper, we attempt to develop a SAM-based zero-shot approach for tracking and segmenting objects in videos without introducing any additional tracking modules.

\section{Method}
\label{sec: Method}
\begin{figure*}[t]
  \centering
  \includegraphics[width=\linewidth]{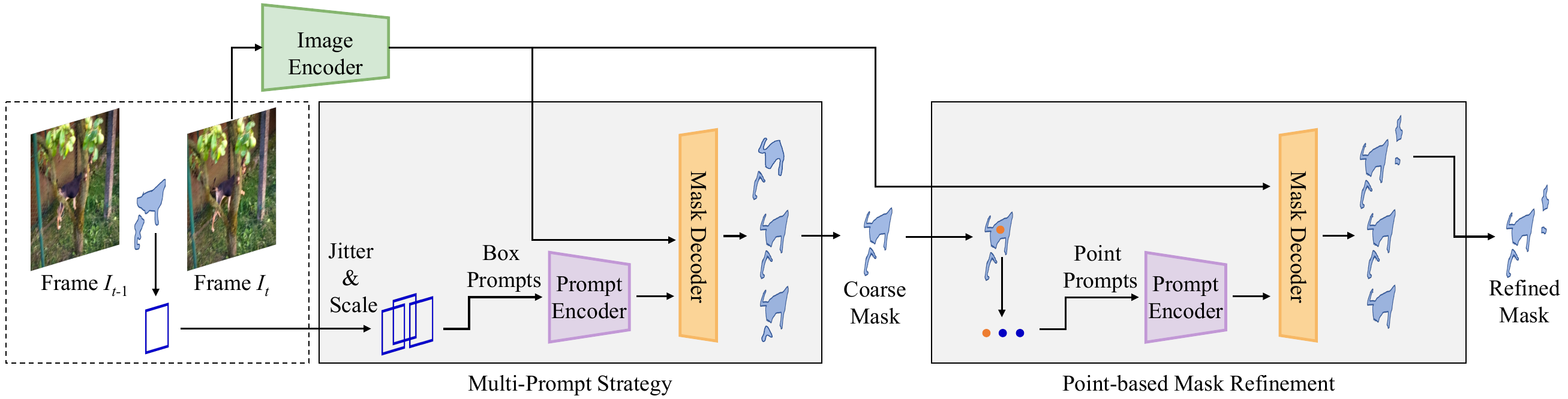}
  \caption{Overview of our method. Given the mask prediction from the previous frame $I_{t-1}$, we propagate its bounding box to the next frame $I_{t}$ as the box prompt for the corresponding object. We augment this vanilla idea with two extensions: a multi-prompt strategy and a point-based mask refinement. The former constructs a group of jittered and scaled box prompts for each object, leading to multiple mask predictions. Among these predictions, we retain the one with the maximum semantic similarity to the template mask (elaborated in Fig.~\ref{fig: multi_box_prompt}). The latter stage takes the coarse mask, samples one positive prompt (orange points) inside it, and combines it with negative point prompts (blue points) sampled from other foreground objects. We refine the coarse mask with these point prompts, leveraging SAM's multi-prediction capability. The two extensions help reduce cumulative errors. For further details, please refer to the document.}
  \label{fig: fig_2}
\end{figure*}

In this section, we introduce the SAM-PD, a SAM-based online approach for zero-shot video object segmentation, where tracking is achieved by prompt denoising. We first provide a brief overview of SAM~\cite{SAM_iccv23} in Sec.~\ref{sec: Preliminary}. Subsequently, we elaborate on our design, including our vanilla idea (in Sec.~\ref{sec: Vanilla Idea}) and two extensions (in Sec.~\ref{sec: Multi-Prompt Strategy} and Sec.~\ref{sec: mask refinement}). Fig.~\ref{fig: fig_2} shows an overview of our method.

\subsection{Preliminary}
\label{sec: Preliminary}
The released version of SAM~\cite{SAM_iccv23} takes positional prompts, including sparse ones
(points, boxes) and dense ones (masks), to segment objects in images. In addition to mask predictions, SAM also outputs a predicted intersection-over-union (IoU) score for each mask to indicate confidence. Driven by its powerful data engine, SAM sets a revolutionary improvement in zero-shot generalization capability. On the model side, SAM adopts a ViT~\cite{ViT_iclr21}-based image encoder, a lightweight prompt encoder, and a lightweight mask decoder. The image encoder accounts for the major computational overhead.

\begin{figure}[t]
  \centering
  \includegraphics[width=\linewidth]{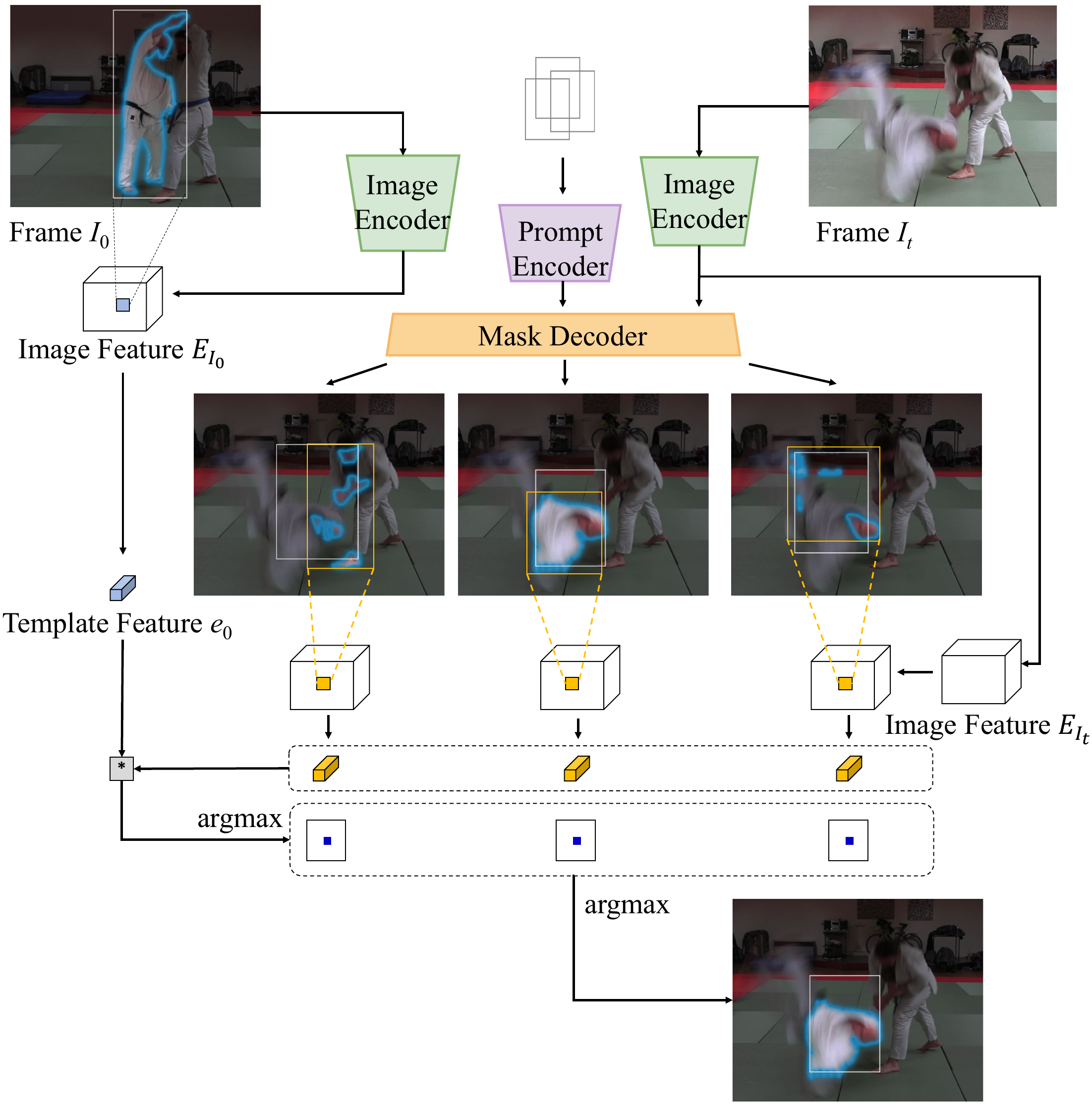}
  \caption{Detailed design of the multi-prompt strategy, where we use white boxes and orange boxes to indicate box prompt inputs and the bounding boxes of mask predictions, respectively.}
  \label{fig: multi_box_prompt}
\end{figure}

SAM supports both an interactive mode and an automatic mode. The interactive mode absorbs user inputs, while the automatic mode automatically segments the full image by evenly placing point prompts on a grid. To resolve ambiguity, when provided with a single-point prompt, SAM outputs three predicted masks, typically corresponding to three layers of the object: whole, part, and subpart.

Considering potentially inaccurate prompt inputs from users (e.g., a loose box), SAM introduces random noise in each coordinate (only for box prompts) during the training phase. The standard deviation is set to $10\%$ of the box side length, with a maximum of $20$ pixels. This allows SAM to exhibit tolerance towards box prompts with minor noise, as observed in Fig.~\ref{fig: fig_1}.

\subsection{Vanilla Idea of SAM-PD}
\label{sec: Vanilla Idea}
Motivated by the observation mentioned above, we explore the potential of using SAM to track and segment objects in videos, where we consider the tracking task as a prompt denoising task. Our approach, named SAM-PD, works in an online manner. For simplicity, we first consider scenarios including a single object. Given the preceding frame $I_{t-1} \in \mathbb{R}^{W \times H \times 3}$ and the predicted mask $m_{t-1} \in \mathbb{R}^{W \times H \times 1}$ for the object of interest, we first extract the bounding box $b_{t-1}$ of the mask $m_{t-1}$. Tracking is achieved by propagating the bounding box $b_{t-1}$ to the next frame $I_{t}$, serving as the box prompt for the corresponding object. Note that, we propagate boxes instead of points because the latter are considered precise (without user noise) by SAM. However, as we will introduce below, leveraging point prompts for refining masks is crucial for handling occlusions, eliminating multi-object ambiguity, and reducing cumulative errors.

\subsection{Multi-Prompt Strategy}
\label{sec: Multi-Prompt Strategy}

Intrinsically, our vanilla method assumes the spatial continuity of objects between adjacent frames. Despite its effectiveness in simple cases, it experiences reduced effectiveness in handling objects with larger displacement or scale changes, as illustrated in Fig.~\ref{fig: multi_box_prompt}.

An intuitive method for displacement compensation is, inspired by SiamFC~\cite{SiamFC_eccv16}, to calculate the semantic correlation heatmap between the template feature and the feature of the current frame, and finally extract the spatial maximum value from the resulting heatmap. For simplicity, we illustrate the case of tracking and segmenting a single object.

Considering that the image encoder accounts for the major computational overhead, we reuse the image features (encoded from the full image) in SAM to save costs. Formally, we adopt following notations: the initial frame image $I_0 \in \mathbb{R}^{W \times H \times 3}$, the current frame image $I_t \in \mathbb{R}^{W \times H \times 3}$, the initial mask\footnote{Each element in $m$ is binary, with 1 representing foreground and 0 representing background. Facilitated by promptable segmentation models, the initial mask $m_{0}$ required by the VOS task can be annotated by providing point or box prompts.} $m_0 \in \mathbb{R}^{W \times H \times 1}$, the SAM image encoder $D_I(\cdot)$, and the encoded image features $E_{I_0} = D_I(I_0)$ and $E_{I_t} = D_I(I_t)$, where $E_{I_0}$ and $E_{I_t} \in \mathbb{R}^{w \times h \times c}$. We use $c$ to denote the dimension of image features. 

We obtain the template feature $e_0$ of the tracked object by cropping the initial image feature $E_{I_0}$ according to the bounding box $b_0$ of the initial mask $m_0$. We filter out background information by applying pixel-wise multiplication between $e_0$ and  $m_0$, resulting in the filtered template feature $\hat{e}_0$. Finally, we calculate the correlation between $\hat{e}_0$ and $E_{I_t}$\footnote{SiamFC~\cite{SiamFC_eccv16} fixes the object feature obtained from the initial frame as the template rather than updating it during tracking, as the latter receives little benefit and the former exhibits better long-term tracking capability. In our experiments, we also tried an updating version but received little benefit.}
, leading to a single-channel heatmap $h_t \in \mathbb{R}^{w' \times h' \times 1}$. This process can be summarized as:
\begin{equation}
  h_t = (e_0 \odot m_0) \ast E_{I_t},
  \label{eq: get_heatmap}
\end{equation}
where $\odot$ denotes the element-wise multiplication, $\ast$ is the correlation operation. The location of the box prompt in the new frame $I_t$ can be compensated by locating the maximum value on the heatmap $h_t$. 

However, we find that such operations do not bring robust position compensation. We attribute this to the positional information (introduced by positional embeddings) inside the image feature, which makes the semantic less discriminative. Another weakness is that it only compensates for the position, without perceiving changes in box size between adjacent frames. 

As an alternative, we introduce a multi-prompt strategy, as illustrated in Fig.~\ref{fig: multi_box_prompt}. This consists of two steps. 

\noindent\textbf{One-to-Many.} For each object, in addition to the original box prompt $b_{t-1}$ propagated from the preceding frame $I_{t-1}$, we construct a prompt group consisting of $N$ jittered and scaled box prompt duplications $\{\tilde{b}^i_t | i=1,\cdots,N\}$. These duplications enrich the coverage in spatial and scale dimensions and result in multiple mask predictions $\{\tilde{m}^i_t | i=1,\cdots,N\}$. We observe that SAM's robustness to noisy box prompts is not isotropic. Minor changes in box prompts may lead to significantly different mask predictions. By constructing multiple box prompts for a single object, this randomness can be reduced, increasing the probability of obtaining relatively high-quality mask predictions. Intuitively, such a multi-prompt strategy is akin to the Monte Carlo search in particle filtering~\cite{particle_filter}, a commonly used non-deep learning method for particle tracking, such as tracking underwater targets or aerial targets.

\noindent\textbf{Many-to-One.} Given the group of mask predictions and their corresponding bounding boxes $\{b^i_t | i=1,...,N\}$, 
we obtain feature crops from the current image feature $E_{I_t}$ and filter out background pixels as mentioned above. Subsequently, we perform correlations between these filtered feature crops and the template feature. This process can be summarized as:
\begin{equation}
  \tilde{h}^{i}_t = (e_0 \odot m_0) \ast (e^{i}_t \odot m^{i}_t). 
  \label{eq: get_heatmap_pro}
\end{equation}
Finally, we perform the $\arg\max$ operation twice, one for extracting the spatially maximum semantic similarity on each heatmap $\tilde{h}^{i}_t$, and the other for preserving the mask prediction $m^{i*}_t$ with the highest semantic similarity.

In cases of tracking multiple objects, we repeat the above operations for each object. Since the prompt encoder and mask decoder are both lightweight, our approach does not introduce significant computational overhead, as analyzed in our experiments.

\subsection{Mask Refinement} 
\label{sec: mask refinement}
To prevent tracking from drifting to background regions or other foreground regions, the multiple box prompts introduced above should be placed near the original one with reasonable sizes (e.g., up to a 10\% difference in both position and scale). Considering SAM is observed to face challenges in recovering a complete mask from over-tight box prompts where SAM tends to predict the area enveloped by the box prompt, this leads to cumulative errors, especially when objects are partially occluded. 

To address this challenge, we append a point-based mask refinement stage. The key insight is to utilize SAM's multiple prediction capability to recover complete masks covering all visible regions of objects. Specifically, we sample one positive point $p_t$ for each object from its coarse mask prediction $m^{i*}_t$. The point is deterministically selected as the pixel farthest from the object boundary, following the point sampling strategy adopted by SAM during the training phase. Given multiple (i.e., 3) mask predictions of each object, we preserve the mask with the largest area as the refined version. However, when required to track a part of an object (e.g., usually required by YouTubeVOS2018~\cite{YouTube2018_dataset}), we retain the mask prediction with the maximum intersection-over-union (IoU) with the mask in the previous frame. We adopt this setting throughout the evaluation on the YouTubeVOS2018~\cite{YouTube2018_dataset} dataset. 

To distinguish between multiple targets, we append points sampled from other objects as negative prompts behind the current object's positive point prompt. Again, due to the lightweight mask decoder, this refinement step involves minor computational costs.

\section{Experiments}
\label{sec: Experiments}
\subsection{Experiment Settings}
\noindent\textbf{Datasets.} We evaluate our method on two tasks: video object segmentation (VOS) and video instance segmentation (VIS). For the evaluation of 
our method on the VOS task, we adopt the commonly used DAVIS 2017~\cite{DAVIS_dataset_arxiv17} and YouTube-VOS 2018~\cite{YouTube2018_dataset} datasets. DAVIS 2017 consists of a total of $150$ video sequences, with $60$ videos for training, $30$ videos for validation, and the remaining for testing. YouTube-VOS 2018 is a large-scale dataset, containing $4,453$ YouTube video clips and $94$ object categories. It includes a validation set with $474$ videos containing $26$ unique categories absent in the training set. So it can be used to evaluate the model's generalization capability. For the VIS task, we adopt the Unidentified Video Objects (UVO) 1.0~\cite{UVO_dataset_ICCV21} dataset, which is constructed for open-world class-agnostic object segmentation in videos. UVO 1.0 has seven times more mask annotations per video compared with YouTube-VOS, and the annotations are not restricted to a close set of objects.

\begin{table*}[t]
    \caption{Results on the validation sets of DAVIS 2017~\cite{DAVIS_dataset_arxiv17} and YouTube-VOS 2018~\cite{YouTube2018_dataset}. *: results reproduced by~\cite{SAM_PT_arxiv23}.}
    \label{table: compare with sota methods}
    \begin{center}
    \begin{small}
    \begin{tabular}{lccccclllll}
    \toprule
           & & & \multicolumn{3}{c}{DAIVS 2017 Validation} & \multicolumn{5}{c}{YouTube-VOS 2018 Validation} \\
    Method & Venue & Propagation & $\mathcal{J\&F}$ & $\mathcal{J}$ & $\mathcal{F}$ & $\mathcal{G}$ & $\mathcal{J}_s$ & $\mathcal{F}_s$ & $\mathcal{J}_u$ & $\mathcal{F}_u$ \\
    \midrule
    \multicolumn{3}{l}{trained on video segmentation data} \\
    \cline{1-11}
    \textcolor{gray}{STM} \cite{STM_iccv19} & \textcolor{gray}{ICCV' 19}  & \textcolor{gray}{-} & \textcolor{gray}{81.8}  & \textcolor{gray}{79.2}  & \textcolor{gray}{84.3}   & \textcolor{gray}{79.4} & \textcolor{gray}{79.7} & \textcolor{gray}{84.2} & \textcolor{gray}{72.8} &  \textcolor{gray}{80.9} \\ 
    \textcolor{gray}{DeAOT} \cite{DeAOT_nips22} & \textcolor{gray}{NIPS' 22} & \textcolor{gray}{-} &  \textcolor{gray}{86.2} & \textcolor{gray}{83.1}  & \textcolor{gray}{89.2} & \textcolor{gray}{86.2} & \textcolor{gray}{85.6} & \textcolor{gray}{90.6} & \textcolor{gray}{80.0} & \textcolor{gray}{88.4}   \\ 
    \textcolor{gray}{XMem} \cite{Xmem_eccv22} & \textcolor{gray}{ECCV' 22}  & \textcolor{gray}{-} & \textcolor{gray}{87.7}  & \textcolor{gray}{84.0}  & \textcolor{gray}{91.4} & \textcolor{gray}{86.1} & \textcolor{gray}{85.1} & \textcolor{gray}{89.8} & \textcolor{gray}{80.3} & \textcolor{gray}{89.2} \\
    \midrule
    \multicolumn{3}{l}{not trained on video segmentation data (zero-shot)} \\
    \cline{1-11}
    Painter \cite{Painter_cvpr23}    &  CVPR' 23 & None & 34.6  & 28.5  & 40.8 & 24.1 & 27.6 & 35.8 & 14.3 & 18.7 \\ 
    SEEM \cite{SEEM_nips23}          &  NIPS' 23 & None & 62.8  & 59.5  & 66.2 & 53.8 & 60.0 & 63.5 & 44.5 & 47.2 \\
    PerSAM\cite{PerSAM_F_arxiv23}&  ICLR' 24 & None & 66.9  & 71.3  & 75.1 & - & - & - & - & - \\
    PerSAM-F\cite{PerSAM_F_arxiv23} (one-shot)&  ICLR' 24 & None & 76.1  & 74.9  & 79.7 & 54.4* & 53.9* & 56.4* & 50.7* & 56.6* \\
    SAM-PT \cite{SAM_PT_arxiv23}     &  arXiv' 23 & Point Tracker & 76.6  & 74.4  & 78.9 & 67.5 & 69.0 & 69.9 & 63.2 & 67.8 \\
    Ours                            &  this work & None & 71.3  & 68.6  & 74.2 & 63.0 & 65.7 & 67.5 & 56.8 & 61.8 \\
    \bottomrule
    \end{tabular}
    \end{small}
    \end{center}
\end{table*}

\vskip 0.15cm
\noindent\textbf{Evaluation Metrics.} 
For the VOS task, we adopt standard VOS metrics~\cite{DAVIS_dataset_arxiv17} including region similarity $\mathcal{J}$, contour accuracy $\mathcal{F}$, and their average $\mathcal{J} \& \mathcal{F}$. Following the common practice~\cite{SAM_PT_arxiv23, Xmem_eccv22, DeAOT_nips22}, we report separate results for ``seen'' and ``unseen'' categories when evaluating with the YouTube-VOS 2018~\cite{YouTube2018_dataset} dataset. For the VIS task, we report the Average Recall (AR) and Average Precision (AP) following~\cite{UVO_dataset_ICCV21, SAM_PT_arxiv23}, which are calculated based on the spatial-temporal intersect-over-union (IoU) between the predicted masks and ground-truth masks.

\vskip 0.15cm
\noindent\textbf{Implementation Details.} We adopt the official SAM checkpoint (the ViT-H version) provided by the authors. The input resolution of the images is $1024 \times 1024$. In our multi-prompt design, $18$ prompts are placed on a $3 \times 3$ grid centered at the original box prompt with a step of $10\%$ of the side length. Two box prompts are placed at each grid with scaling factors of $1$ and $1.05$, respectively. We elaborate on different number settings in Sec.~\ref{sec: Ablation Study}. The time costs below are tested using a single NVIDIA RTX 3090 GPU.

\begin{table}[t]
    \caption{Comparative results on the UVO 1.0~\cite{UVO_dataset_ICCV21} validation set.}
    \label{table: uvo results}
    \begin{center}
    \begin{small}
    \resizebox{\linewidth}{!}{ 
    \begin{tabular}{cccccc}
    \toprule
    Method &  AR100 & ARs & ARm & ARl & AP \\
    \midrule
    \textcolor{gray}{Mask2Former VIS} \cite{Mask2Former_VIS_arxiv22} & \textcolor{gray}{35.4} & \textcolor{gray}{-} & \textcolor{gray}{-} & \textcolor{gray}{-} & \textcolor{gray}{27.3} \\ 
    \midrule
    TAM \cite{TAM_arxiv23}             & 24.1 & 21.1 & 32.9 & 31.1  & 1.7 \\ 
    SAM-PT \cite{SAM_PT_arxiv23}       & 30.8 & 25.1 & 44.1 & 49.2 & 6.5 \\
    Ours                              & 28.1 & 22.6 & 40.7 & 46.0 & 6.4 \\
    \bottomrule
    \end{tabular}}
    \end{small}
    \end{center}
\end{table}

\subsection{Main Results}
\label{sec: Main Results}
\noindent\textbf{Video Object Segmentation.} 
In the VOS task, the ground truth masks are provided in the initial frame and the model is required to propagate them throughout the video. In Table~\ref{table: compare with sota methods}, we present several supervised methods and concurrent studies that share similar zero-shot settings as ours. Specifically, Painter~\cite{Painter_cvpr23} presents a general vision model that performs vision tasks according to the input task prompts, eliminating the need for task-specific heads. It is achieved by performing standard masked image modeling on the stitch of input and output image pairs. Despite generalist, Painter shows limited effectiveness in the VOS task. SEEM~\cite{SEEM_nips23} is a concurrent work to SAM. Compared to SAM, SEEM additionally supports image segmentation by providing a reference region from another image. This capability allows SEEM to be naturally applied to the VOS task. However, on DAVIS 2017 and YouTube-VOS 2018 datasets, our approach outperforms SEEM by a large margin. We attribute this to the distinct ability to distinguish multiple visually similar objects. PerSAM-F~\cite{PerSAM_F_arxiv23} is another recent work of reference image segmentation. It utilizes semantics inside SAM feature maps to locate the approximate position of the reference object in the query image. Our method outperforms the zero-shot version of PerSAM-F (denoted as ``PerSAM'' in Table~\ref{table: compare with sota methods}). While its one-shot version achieves better performance on the DAVIS 2017 dataset, it still exhibits limited performance on the YouTube-VOS 2018 dataset. SAM-PT uses point trackers~\cite{PIPS_eccv22, Tap_Vid_nips22} to propagate points sampled from the initial mask to subsequent frames, serving as point prompts for SAM to obtain mask predictions. Driven by the powerful point tracker, SAM-PT exhibits better robustness to variations in object position, size, and visibility. In general, our approach demonstrates comparable performance, especially among methods that do not introduce additional tracking modules.

\vskip 0.2cm
\noindent\textbf{Video Instance Segmentation.}
The VIS task requires the model to detect, track, and segment each instance throughout the video sequence. 
When evaluating our method on the VIS task, we first generate up to $100$ masks in the initial frame by switching the SAM to the automatic mask generation mode. We then propagate these initial masks throughout the video without spawning new mask trajectories in subsequent frames. Our experimental settings are the same as those of SAM-PT~\cite{SAM_PT_arxiv23}. Shown in Table~\ref{table: uvo results}, despite not being trained on video segmentation data, SAM-PD benefits from the strong zero-shot generalization capability of SAM, surpassing TAM~\cite{TAM_arxiv23} on the UVO 1.0 dataset. Besides, without introducing additional tracking models, our method demonstrates comparable performance compared to SAM-PT~\cite{SAM_PT_arxiv23}.  

\subsection{Ablation Study}
\label{sec: Ablation Study}
In this section, we ablate our design, including the vanilla idea, the multi-prompt strategy, and the point-based mask refinement. The results are reported in Table~\ref{table: ablation study}. 
\begin{table}[t]
    \caption{Ablation studies on the DAVIS 2017 validation set.}
    \label{table: ablation study}
    \begin{center}
    \begin{small}
    \resizebox{\linewidth}{!}{
    \begin{tabular}{cccccccc}
    \toprule
    Vanilla & Multi-Prompt & Mask Refine & $\mathcal{J\&F}$ & $\mathcal{J}$ & $\mathcal{F}$ \\
    \midrule
    Point &            &            & 33.0 & 31.1 & 35.0 \\
    Point &            & \checkmark & 49.5 & 47.5 & 51.4 \\
    \midrule
    Box   &            &            & 67.2 &  64.4 & 69.9 \\ 
    Box   & \checkmark &            & 69.0 &  66.3 & 71.8 \\ 
    Box   &            & \checkmark & 68.4 &  65.9 & 71.0 \\ 
    Box   & \checkmark & \checkmark & 71.3 &  68.3  & 74.2 \\ 
    \midrule
    Box   & IoU-based & \checkmark & 70.2 &  67.9  & 72.5 \\
    Box   & Semantic-based & \checkmark & 71.3 &  68.3  & 74.2 \\
    \bottomrule
    \end{tabular}}
    \end{small}
    \end{center}
\end{table}

\noindent\textbf{Vanilla Method.} As shown in the third row of Table~\ref{table: ablation study}, without bells and whistles, our vanilla method demonstrates reasonable effectiveness by simply propagating the bounding box of the mask from the previous frame to the next as a box prompt. This is because most objects in video sequences exhibit spatial continuity between adjacent moments, especially in high-frame-rate cases. Additionally, as an alternative to the box prompt, we report the performance of point propagation in the first and second rows of Table~\ref{table: ablation study}. Specifically, we deterministically sample points inside each mask that are furthest from the object boundary, serving as the point prompt for the next frame. The unsatisfactory results stem from the limitations of single-point prompts, namely (1) susceptibility to ambiguity and (2) difficulties in tracking non-convex objects. Exploring advanced point sampling strategies may improve the results.

\noindent\textbf{Multiple Prompts.} 
The multi-prompt strategy is observed to improve the performance of the vanilla method in Table~\ref{table: ablation study} (69.0 \emph{vs.} 67.2 of $\mathcal{J\&F}$ score), and when combined with the mask refinement strategy, it further enhances performance. Further, we explore retaining the mask with maximum IoU similarity instead of the one with maximum semantic similarity during the ``many-to-one'' stage. The former does not lag far behind the latter (70.2 vs. 71.3 of $\mathcal{J\&F}$ score), indicating the effectiveness of spatial cues, and the limited semantic discriminative ability of SAM (discussed later).

In Table~\ref{table: box number}, we report the performance and time cost under different settings of the number of boxes $N$ on the validation set of DAVIS 2017~\cite{DAVIS_dataset_arxiv17}. Except for when using a single box prompt (where semantic similarity computation is not required), the time cost increases linearly with the number of prompts. We experimentally set $N_{box}=9$ and $N_{scale}=2$ by default. Compared to the heavy image encoding overhead, the time cost introduced by the multi-prompt strategy is acceptable. In the future, our approach can be expected to benefit from more efficient variants of SAM, such as MobileSAM~\cite{MobileSAM_Arxiv23} and EfficientSAM~\cite{EfficientSAM_Arxiv23}.

\noindent\textbf{Mask Refinement.} As observed in the $5^{\text{th}}$ and $6^{\text{th}}$ rows of Table~\ref{table: ablation study}, engaging point-based mask refinement helps overcome incomplete mask predictions caused by over-tight box prompts, reducing cumulative errors.

\noindent\textbf{Negative Point Prompt.} Finally, in Table~\ref{table: negative point query}, we investigate the effectiveness of using points sampled from other objects as negative point prompts for the current object. It is observed to statistically bring improvements in both the multi-prompt stage and the mask refinement stage, indicating its utility in aiding the model to distinguish between different objects. However, using negative point prompts may occasionally lead to unstable mask predictions, especially during the box prompt stage.

\begin{table}[t]
    \caption{Time cost (on single Nvidia RTX3090) of the multi-prompt stage under different box number settings.}
    \label{table: box number}
    \begin{center}
    \begin{small}
    \begin{tabular}{cccccccc}
    \toprule
    $N_{box}$ & $N_{scale}$ & Time Cost (ms) & $\mathcal{J\&F}$ & $\mathcal{J}$ & $\mathcal{F}$ \\
    \midrule
    \multicolumn{5}{l}{SAM-ViT-H (image encoding costs 422 ms)} \\
    \midrule
    1     & 1             &   6.1    & 68.4 &  65.9 & 71.0 \\ 
    $3^2$   & 1           &   84.2   & 70.7 &  67.9 & 73.4 \\ %
    $3^2$   & 2           &   168.8   & 71.3 &  68.3 & 74.2  \\ %
    $5^2$   & 2           &   415.3   & 68.7 &  66.0 & 71.4 \\ 
    \bottomrule
    \end{tabular}
    \end{small}
    \end{center}
\end{table}

\begin{table}[t]
    \caption{Effectiveness of negative point queries.}
    \label{table: negative point query}
    \begin{center}
    \begin{small}
    \begin{tabular}{ccccc}
    \toprule
    \multicolumn{2}{c}{Use Negative Point Prompt} &  & &  \\
    Multi-Prompt  & Mask Refine & $\mathcal{J\&F}$ & $\mathcal{J}$ & $\mathcal{F}$ \\
    \midrule
    \checkmark &             &  70.4  & 67.0 & 73.8  \\ 
               & \checkmark  &  70.4  & 67.5 & 73.4 \\ 
    \checkmark & \checkmark  &  71.3 & 68.3 & 74.2 \\ 
    \bottomrule
    \end{tabular}
    \vskip -0.5cm
    \end{small}
    \end{center}
\end{table}

\begin{figure*}[t]
  \centering
  \includegraphics[width=\linewidth]{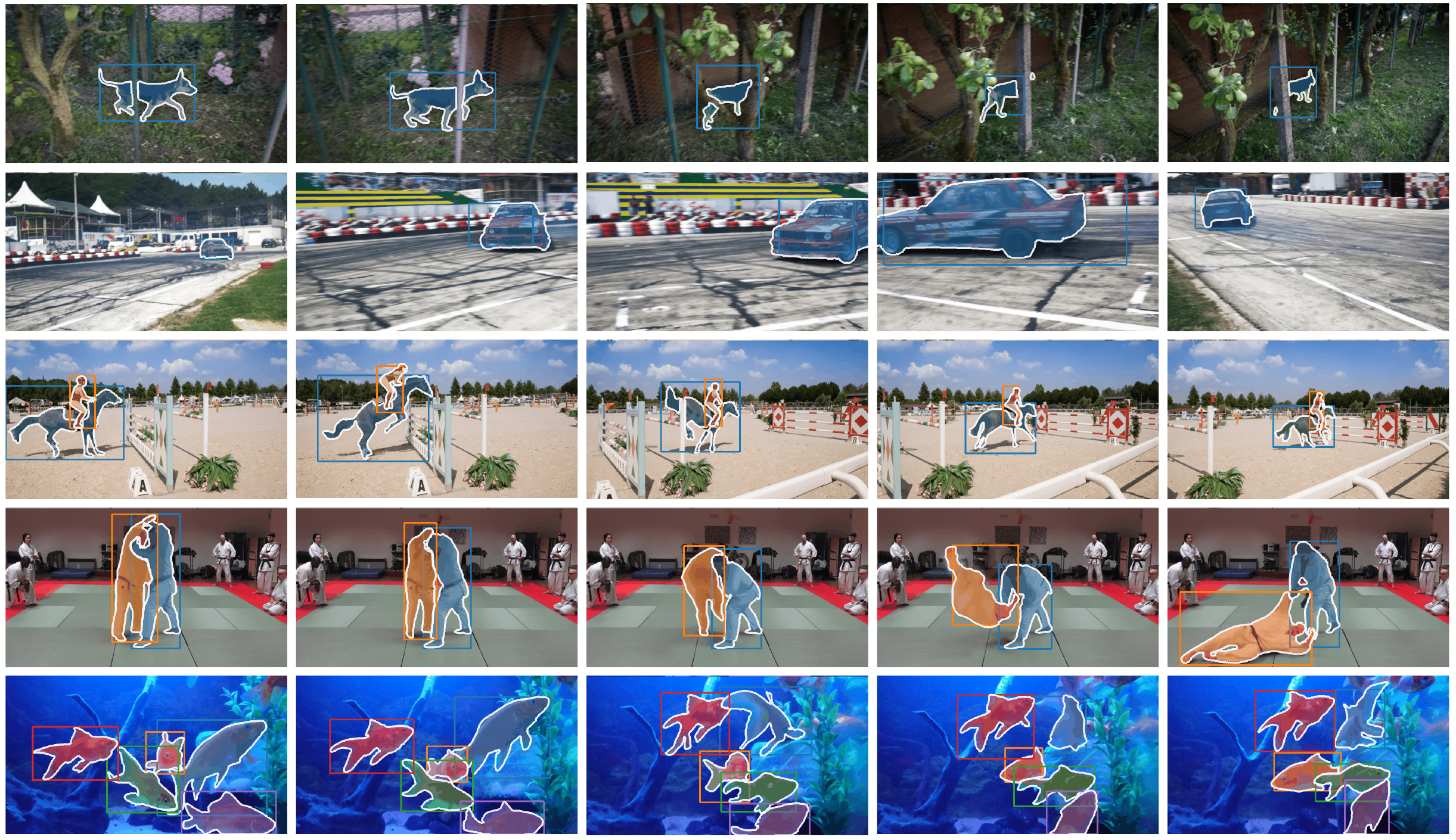}
  \caption{Qualitative results on DAVIS 2017 validation set. For each predicted mask, we also plot the box prompt propagated from the previous frame.}
  \label{fig: qualitative_results_good}
\end{figure*}

\begin{figure*}[t]
  \centering
  \includegraphics[width=\linewidth]{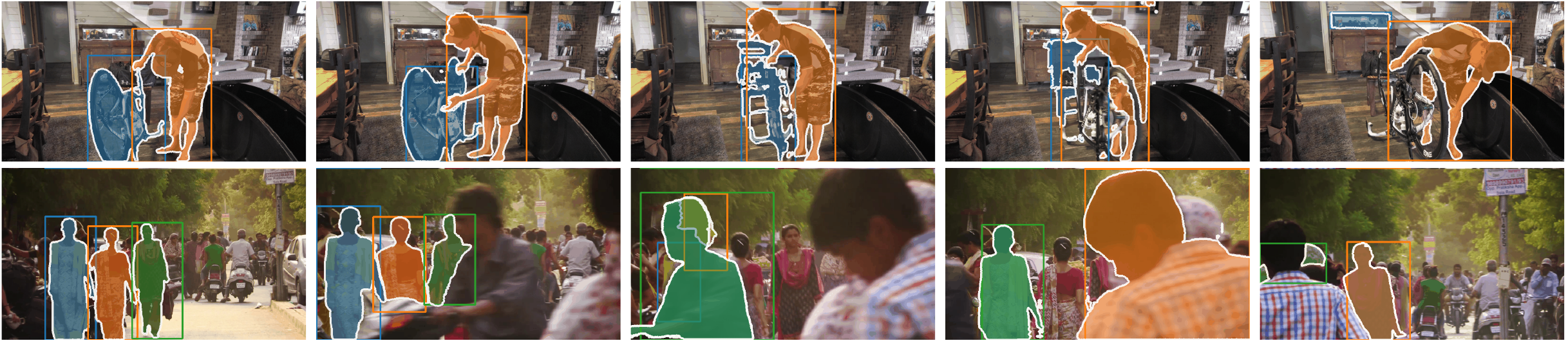}
  \caption{Failure cases. Our method is bounded by the stability of SAM to box prompts (the third frame of the first case) and faces challenges in handling large object motion (the fourth frame of the first case) and full occlusion.}
  \label{fig: qualitative_results_failure}
\end{figure*}

\section{Conclusion}
This paper introduces an online method, named SAM-PD, that applies SAM to track and segment objects throughout the video. By recognizing the tracking task as a prompt denoising task, we iteratively propagate the bounding box of each object's mask in the preceding frame as the prompt for the next. Two extensions, a multi-prompt strategy, and a point-based refinement are proposed to enhance SAM-PD's capability to handle position, size, and visibility changes. The proposed method shows comparable performance on three datasets: DAVIS2017, YouTubeVOS2018, and UVO, without involving external tracking modules. In general, SAM-PD serves as a concise baseline that endows potential SAM-based downstream applications with tracking capabilities.

\noindent\textbf{Limitations and Future Work.} 
SAM-PD tracks objects by propagating spatial prompts and reserves masks based on semantic similarity in the multi-prompt strategy. The effectiveness of the proposed method 
(1) reduces in handling large motion, full occlusion, crowded objects, or discontinuous scenes; and (2) is bounded by the limited semantic distinctiveness of the image features in SAM. We provide further analysis of the latter in the supplement. We notice that another promptable segmentation model, SEEM~\cite{SEEM_nips23}, has achieved better semantic distinctiveness, but still demonstrated limited semantic-based VOS effectiveness in challenging cases. We look forward to future foundation models demonstrating higher-order semantic understanding (e.g., occlusion relationships, relative positions). To further improve the performance of SAM-PD, one may (1) use semantic features with better discrimination~\cite{DINO_iccv21}; (2) adopt a customed version of SAM with better robustness to imprecise box prompts~\cite{stable_sam_arxiv23}; and (3) introduce a memory mechanism to recover tracking identities for completely occluded objects.



{
    \small
    \bibliographystyle{IEEEtran}
    \bibliography{main.bib}
}

\clearpage

\appendix

In the main text, we propose a SAM~\cite{SAM_iccv23}-based online method, named SAM-PD, for tracking and segmenting anything by recognizing the tracking task as a prompt denoising task. This document contains additional material for the main submission. In Sec.~\ref{supsec: Oracle Experiments}, we quantitatively evaluate the prompt denoising capability of SAM through oracle experiments. Sec.~\ref{supsec: Semantic} analyzes the semantics in SAM's latent space by providing detailed semantic similarity scores.

\section{Oracle Experiments of Prompt Denoising}
\label{supsec: Oracle Experiments}

In our main text, the promptable segmentation model, SAM~\cite{SAM_iccv23}, was observed to be robust when given box prompts with minor noise. Such robustness arises from SAM simulating inaccurate user input during the training process, where SAM~\cite{SAM_iccv23} injects random noise into each box prompt, with a standard deviation of 10\% box side length (up to a maximum of $20$ pixels). In this section, we conduct oracle experiments to quantitatively evaluate SAM's denoising capability. 

\begin{figure*}[h]
  \centering
    \begin{subfigure}[h]{\linewidth}
         \centering
         \includegraphics[width=\linewidth]{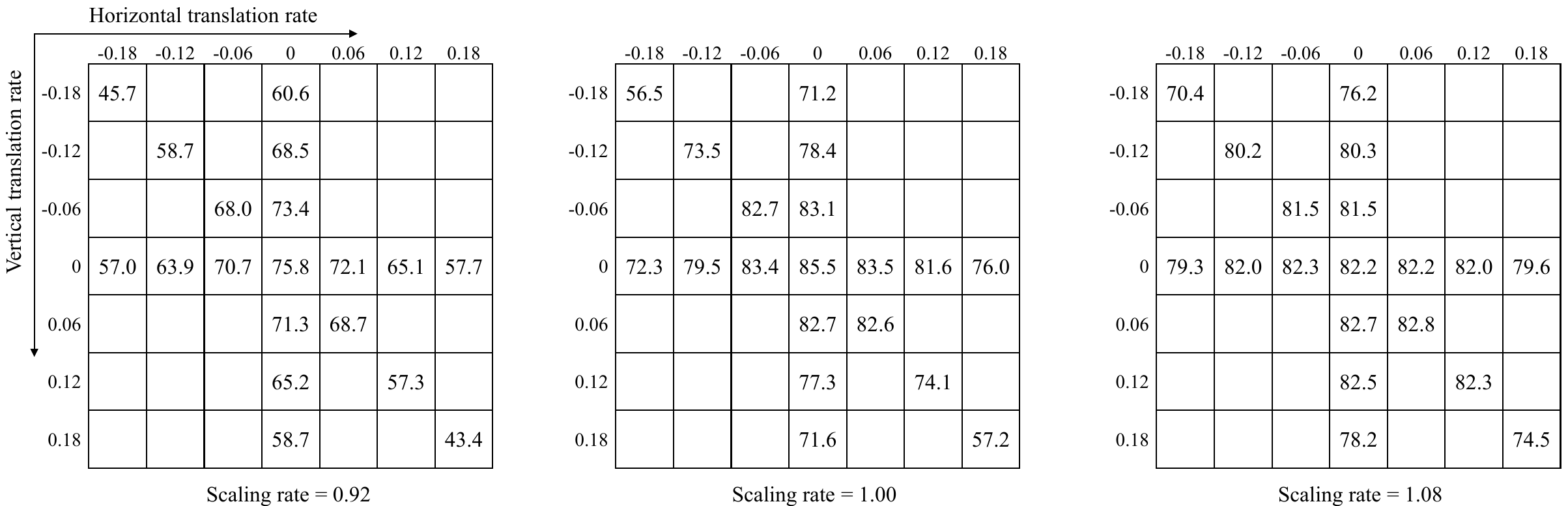}
         \caption{Single box prompt.}
         \label{fig: Oracle_denoise_exp_vanilla}
    \end{subfigure}
    \begin{subfigure}[h]{\linewidth}
         \centering
         \includegraphics[width=\linewidth]{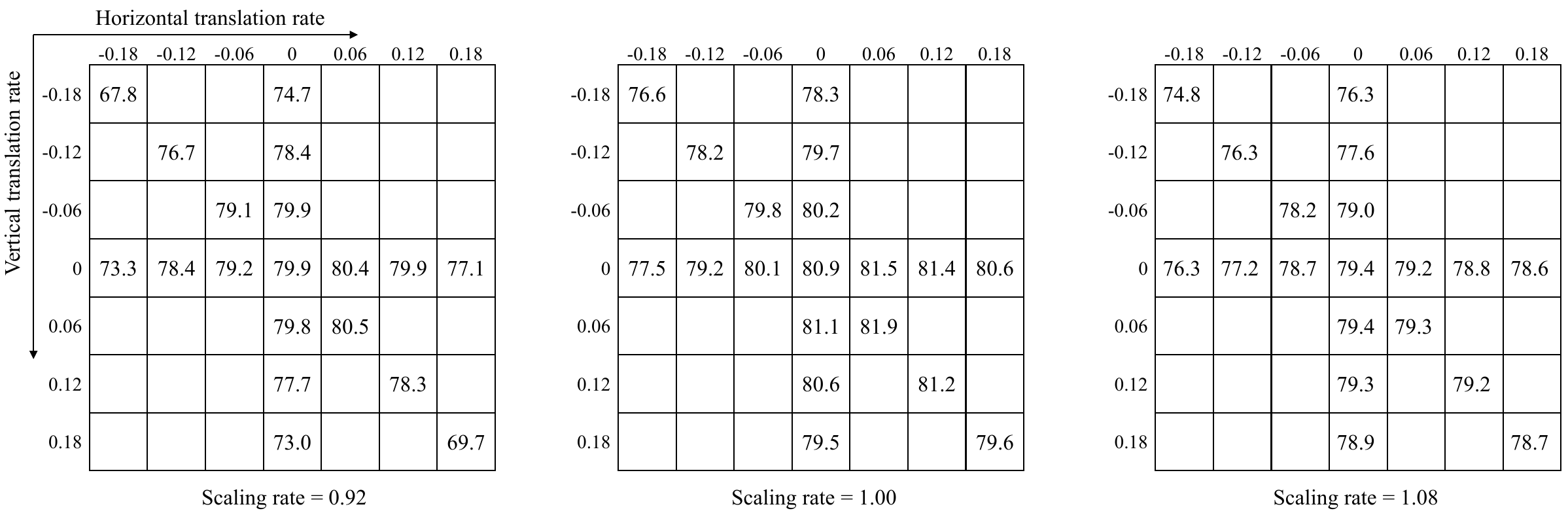}
         \caption{Multiple box prompts.}
         \label{fig: Oracle_denoise_exp_multi_prompt}
    \end{subfigure}
  \caption{Oracle experiments on the DAVIS 2017 validation set. We report the $\mathcal{J\&F}$ score under various translation and scaling rate combinations. Please note that there is no tracking involved. The vanilla box prompts for each frame are the perturbed ground-truth bounding boxes of the objects in that frame.}
  \label{fig: Oracle_denoise_exp}
\end{figure*}

Experiments are conducted on the DAVIS 2017~\cite{DAVIS_dataset_arxiv17} validation set. Specifically, we first extract the bounding box of the ground-truth mask for each object in the \textbf{current frame} and then simulate variations in object position and size by translating and scaling the bounding box. We use the translated and scaled box as the box prompt to query SAM and obtain the predicted mask. Quantitative results are reported in Fig.~\ref{fig: Oracle_denoise_exp}, where we calculate the $\mathcal{J\&F}$ score between the ground-truth mask and the predicted mask. Please note that in this oracle experiment, there is no tracking involved. The box prompts for each frame are the perturbed ground-truth bounding boxes of the objects in that frame. We enumerate various combinations of scaling rates and horizontal/vertical translation rates. Detailed settings are provided in Table~\ref{table: Oracle_exp_settings}. Additionally, we report the performance of incorporating the multi-prompt strategy (introduced in Sec. 3.3 of the main text), with corresponding parameter settings listed in Table~\ref{table: Oracle_exp_settings}. For practical concerns, we make its parameter settings misaligned with the oracle settings.

\begin{figure}[h]
  \centering
  \includegraphics[width=0.9\linewidth]{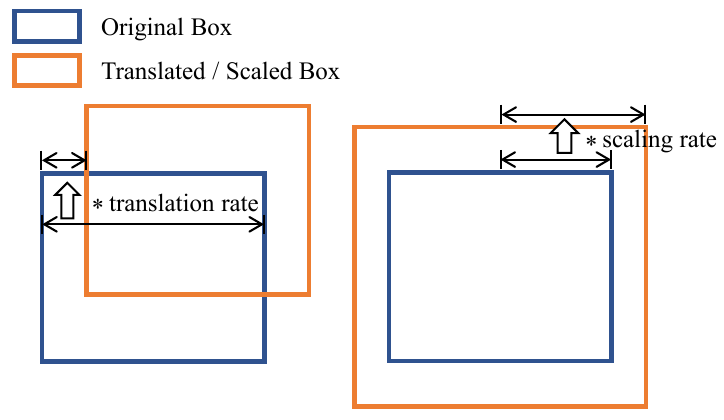}
  \caption{Definitions of translation rate and scaling rate.}
  \label{fig: Shift_and_Scale_box_example}
\end{figure}

\begin{table}[h]
    \caption{Oracle experiment settings.}
    \label{table: Oracle_exp_settings}
    \begin{center}
    \begin{small}
    \vskip -0.5cm
    \resizebox{\linewidth}{!}{
    \begin{tabular}{ccc}
    \toprule
                 & Translation Rate & Scaling Rate \\
    \midrule
    Oracle       &  [-0.18, -0.12, -0.06, 0, 0.06, 0.12, 0.18]   & [0.92, 1.00, 1.08] \\ %
    Multi-Prompt &  [-0.1, 0, 0.1]   &  [0.95, 1.00, 1.05] \\ %
    \bottomrule
    \end{tabular}}
    \end{small}
    \end{center}
\end{table}

\begin{table}[h]
    \caption{Vanilla VOS performance against box prompt scaling rate. Experiments are conducted on the DAVIS 2017 validation set.}
    \label{table: vanilla performance against scaling rate}
    \begin{center}
    \begin{small}
    \vskip -0.5cm
    \resizebox{\linewidth}{!}{
    \begin{tabular}{ccccccccc}
    \toprule
    Scaling rate             & -0.04 & -0.02 & 0 & 0.02 & 0.04 & 0.06 & 0.08 \\
    \midrule
    $\mathcal{J\&F}$  & 43.9 & 61.3 & 65.6 & 65.8 & 65.6 & 65.3 & 64.1 \\ 
    \bottomrule
    \end{tabular}}
    \end{small}
    \end{center}
\end{table}

According to Fig.~\ref{fig: Oracle_denoise_exp}, we have several observations. 

\noindent\textbf{Compared to over-tight boxes, SAM prefers loose boxes.} Shown in Fig.~\ref{fig: Oracle_denoise_exp_vanilla}, 
under each translation setting, the performance with a loose prompt box surpasses that with an over-tight box. On the one hand, this indicates that SAM faces challenges in recovering the complete mask from an over-tight bounding box query, validating the necessity of our multi-prompt strategy and the point-based mask refinement design. On the other hand, however, this does not mean that a loose box prompt is preferred for tracking. To illustrate this point, we evaluate our vanilla method with scaled prompt boxes coming from the previous frame and report the video object segmentation (VOS) performance in Table~\ref{table: vanilla performance against scaling rate}. It can be observed that the optimal performance is achieved around the scaling rate of $1.00$. This is because a loose box prompt may cause SAM to focus on other regions, especially when the tracked object does not occupy the dominant area of the box. 

\noindent\textbf{Effectiveness of the proposed multi-prompt strategy.} 
By comparing Fig.~\ref{fig: Oracle_denoise_exp_vanilla} and Fig.~\ref{fig: Oracle_denoise_exp_multi_prompt}, the multi-prompt strategy is witnessed to significantly improve the denoising capability when the box prompt deviates substantially from the correct object in both spatial and scale aspects. However, it exhibits minor negative effects when the box prompt only contains minor noise. We attribute the latter to two reasons. First, recalling our multi-prompt strategy, it relies on semantic similarity to retain the best coarse mask. As indicated by Eq. (2) in the main text, such semantic similarity is obtained by comparing the latent embedding of the current frame's mask with that of the mask in the initial frame. While this may lead to better long-term tracking performance~\cite{SiamFC_eccv16}, the similarity may diminish as the distance from the initial moment increases. Second, this is also attributed to the limited semantic discrimination in SAM's latent space, as we will analyze in the next section.

\section{Semantics in the Latent Space of SAM}
\label{supsec: Semantic}
\begin{figure*}[h]
  \centering
  \includegraphics[width=\linewidth]{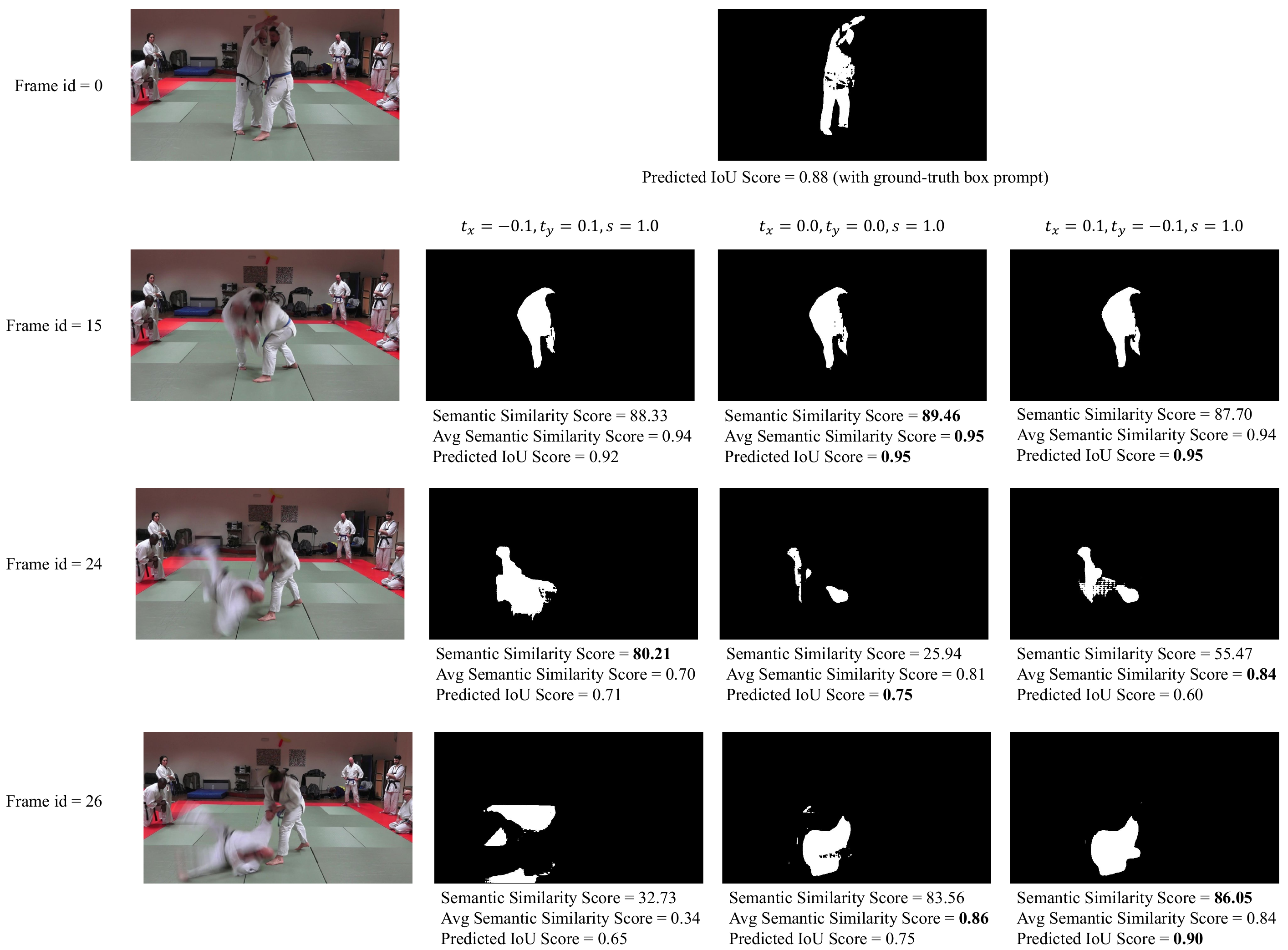}
  \caption{Illustration of semantics in the latent space of SAM. We take the example of tracking and segmenting a single object. For each frame, we visualize the mask predictions corresponding to three prompts. We use $t_x$ and $t_y$ to denote their translation rates in vertical and horizontal directions, and $s$ to represent the scaling rate. Below each mask prediction, we report the semantic similarity score (obtained through Eq. (2) in the main text), the average semantic similarity score (normalized by the area of the predicted mask), and the IoU score predicted by SAM.}
  \label{fig: semantic visualize}
\end{figure*}

In this section, we analyze the semantic discrimination in SAM's latent space. To this end, in Fig.~\ref{fig: semantic visualize}, we visualize the mask predictions corresponding to three prompts in the multi-prompt strategy for each frame.
 Below each mask prediction, we report the semantic similarity score (obtained through Eq. (2) in the main text), the average semantic similarity score (normalized by the area of the predicted mask), and the intersection-over-union (IoU) score predicted by SAM. According to Fig.~\ref{fig: semantic visualize}, we have the following observations.
\begin{itemize}
    \item In general, higher semantic similarity scores correspond to better mask predictions.
    \item However, higher average semantic similarity scores do not necessarily indicate better predictions.
    \item From the results of the 24$^{th}$ frame, a higher IoU prediction score does not necessarily mean better mask predictions.
    \item Indicated by the average semantic similarity score at the 26$^{th}$ frame, it is observed that SAM exhibits semantic discrimination between the background and foreground (0.34 vs. 0.86). It might be more desirable if the semantic similarity between the foreground and background could show a negative value (less than zero), though this might be challenging for the current version of SAM, as it was trained without explicit semantic supervision, and its latent space contains positional information (introduced through positional embedding).
    \item It can be observed that box prompts with different translation rates alternately achieve the maximum semantic similarity. This validates that the proposed multi-prompt strategy helps overcome occasional failures of a single box prompt, enhancing SAM-PD's ability to handle changes in object position and size.

\end{itemize}

\end{document}